\pdfoutput=1

\documentclass[11pt]{article}

\usepackage{emnlp2021}

\usepackage{times}
\usepackage{latexsym}
\usepackage{xurl}
\usepackage[T1]{fontenc}

\usepackage[utf8]{inputenc}

\usepackage{microtype}

\usepackage{mystyle}

\usepackage{graphicx}
\usepackage{amsmath}
\usepackage{multirow}
\usepackage{comment}
\usepackage{appendix}
\usepackage{float}
\usepackage{placeins}
\usepackage{microtype}
\usepackage{wrapfig}
\usepackage{algorithmic}  
\usepackage[algo2e, ruled]{algorithm2e}
\usepackage{booktabs}

\newcommand{\theTitle}{\oscar: Orthogonal Subspace Correction and Rectification of Biases in Word Embeddings}

\title{\theTitle}

\author{Sunipa Dev \\
  UCLA \\
  \And
  Tao Li \\
  University of Utah \\
  \And
  Jeff M Phillips \\
  University of Utah \\
  \And
  Vivek Srikumar \\
  University of Utah 
  
  \AND
  
 \tt{sunipa@cs.ucla.edu \{tli,jeffmp,svivek\}@cs.utah.edu}
 }

\date{}

\begin{document}
\maketitle
\begin{abstract}
Language representations are known to carry certain associations (e.g., gendered connotations) which may lead to invalid and harmful predictions in downstream tasks.
While existing methods are effective at mitigating such unwanted associations by linear projection, we argue that they are too aggressive:
not only do they remove such associations, they also erase information that should be retained.
To address this issue, we propose \oscar (Orthogonal Subspace Correction and Rectification), a balanced approach of mitigation that focuses on disentangling associations between concepts that are deemed problematic, instead of removing concepts wholesale.
We develop new measurements for evaluating information retention relevant to the debiasing goal.
Our experiments on gender-occupation associations show that \oscar is a well-balanced approach that ensures that semantic information is retained in the embeddings and unwanted associations are also effectively mitigated.
\end{abstract}

\section{Introduction} \label{sec:intro}
Word embeddings are used extensively across natural language processing (NLP) 
and succinctly capture not only the syntactic and semantic structure  
of language, but 
also word meaning in context.
As such, word embeddings are essential building blocks for today's state-of-the-art in NLP. 
But they are also known to capture a significant amount of stereotypical associations \cite[e.g.,][]{debias, zhao-etal-2017-men, Bias1, biasSurvey} related to gender, race, nationality, or religion, which can manifest in unwanted and/or potentially harmful ways in downstream tasks \cite{gap, zhao-etal-2019-gender,bias2}. Such potentially problematic associations, when embedded in word representations, can lead to incorrect and unfair decisions about large groups of people.  While the term ``\emph{bias}'' has many meanings, in this paper we use it to refer to these {\em unwanted stereotypical associations}. 

Existing methods to mitigate these effects either require expensive retraining of vectors \cite{gan-bias} which can be inefficient, or projecting out information contained along an entire subspace representing a protected concept (such as gender or race) in the embedding space \cite[e.g.,][]{debias,Bias1, ravfogel2020null}.
Projective approaches are difficult to control as they are either insufficient: removing a subspace can still leave residual bias~\cite{gonen2019lipstick, lauscher2019bias}, or too aggressive: in the case of gender, also unnecessarily altering the association between the word \emph{pregnant} and words like \emph{female} and \emph{mother}.
In  tasks such as coreference resolution, removing such associations could  hinder reference resolution.  

To quantify how much valid information is retained, we look at the output space of models.
Following~\citet{bias2}, we use Natural Language Inference (NLI) as an effective quantitative probe.
Here, we construct the hypothesis by making minimal edits to a premise, and observe model prediction conditioned on such changes. For example,
\begin{description}[noitemsep,topsep=0.5ex,align=right,labelindent=12ex]
\item [Premise:] A \subject{matriarch} sold a watch.
\item [Hypothesis:] A \subject{woman} sold a watch.
\end{description}

Here, the objective is to determine if the hypothesis is \emph{entailed} by the premise, \emph{contradicted} by it, or neither (\emph{neutral} to it). A GloVe-based NLI~\cite{parikh2016decomposable} model, without any explicit form of bias mitigation, predicts label \emph{entail} with a high probability of $97\%$; the notion of a \subject{matriarch} being a \subject{woman} is correctly identified by the model.  
However, after projective debiasing, 
the model classifies the pair as  \emph{neutral} with a probability $62\%$ while the probability of the label $\emph{entail}$ drops to much lower at $16\%$. 
That is, \emph{aggressive mitigation of gender representations erases valid gender associations.}




Ideally, we should correct problematic associations without erasing valid ones.
To this end, we propose \oscar (Othogonal Subspace Correction and Rectification) which \emph{orthogonalizes} and \emph{rectifies} identified subspaces of concepts that are incorrectly associated in an embedding space.
Embeddings outside the subspace are stretched in a graded manner or untouched.
Our contributions are:
\begin{enumerate} 
\item We argue that mitigating unwanted stereotypical associations should go beyond information removal (e.g., projecting out features), and should also preserve pertinent associations.  
\item We present \oscar\footnote{\url{https://github.com/sunipa/OSCaR-Orthogonal-Subspace-Correction-and-Rectification/tree/transformer}}, a completely different method from the existing projective approaches;  it uses orthogonalization of subspaces desired not to have interdependence, and so minimal change is made to embeddings to prevent loss of desired associations.  
\item We develop a combination of tests based on NLI to evaluate both bias mitigation and information retention.
\item Our experiments show that \oscar is a well-balanced approach that mitigates biases as good as projective approaches while retaining more valid associations.
\end{enumerate}

Our contributions, focusing specifically on representations rather than classifiers, are important given the preponderance of distributed representations of text across NLP.
Predictions from systems that use such representations, if unchecked, could lead to real-world decisions (involving, e.g. hiring) that ``\emph{systematically} and \emph{unfairly} discriminate against certain individuals or groups of individuals in favor of others''~\cite{friedman1996bias}.
In an effort to prevent such transition of representational harms into allocational harms~\cite[cf.][]{crawford2017trouble,abbasi2019fairness,blodgett-etal-2020-language}, we look at mitigating stereotyping biases at the source, i.e. the embedding space. 

\section{Bias, Gender and NLP}
\paragraph{Bias and Related Work.} 
Social biases in tasks using machine learning have the potential to cause harms~\cite{barocas2016big} and are widely studied.
Such biases in language technologies have been detected~\cite{debias,gonen2019lipstick}, measured~\cite{Caliskan183,gap,lauscher2019bias} and mitigated~\cite{gn-glove,ravfogel2020null}. Most work focuses on gender bias, and in particular, the stereotypical associations of occupations to males and females~\cite{rudinger2018gender,De-artega2019bias,gaut2019towards,bias2}.
In this work too, bias refers to stereotypical associations in language representations.

\paragraph{Treatment of Gender.}
The concept of gender is a complex one~\cite[cf.][]{larson-2017-gender,dev2021harms}. In our experiments, we identify a subspace in the representation associated with gender, typically derived by terms specifically associated with notions of male and female.  This aligns with prominent recent NLP research~\cite{Caliskan183,debias,Bias1,ravfogel2020null} that try to measure and mitigate gender bias (either implicitly or explicitly), directly focusing on male versus female associations.

The underlying goal of this work is to disassociate two concepts.  We mainly use concepts male-vs-female and occupations to demonstrate this.  There is explicit and measurable bias encoded in word vector representations~\cite{debias} by the male-vs-female relation to occupations, and standard evaluation metrics~\cite{Caliskan183,bias2} are designed to measure this.  As such many of our experiments are geared towards these male-vs-female issues.  

The notion of gender is not restricted to male and female.  But because
``conceptualizations of gender as binary are reinforced linguistically in English through pronoun conventions''~\cite{blodgett2021sociolinguistically}, gender notions beyond male and female are either missing or poorly represented.  
As a result, their associations with other concepts are also not robust. 
Quantifying this effect, and evaluating its attenuation, is beyond the scope of this paper; however the techniques we study for disassociating two concepts are amenable to such tasks in general.  
We remark that a potential harm towards non-male/female people arises through omission~\cite{cao2020towards}, and as a result, we explicitly encourage further research on this topic.

\section{Quantifying Bias and Information Retained}
Keeping in mind that removing bias and retaining information have to be done in synergy,
we present how to obtain aggregated measurements for these two components. 
We will first describe the design of our evaluation dataset in \S\ref{sec:data-design},
then present evaluation metrics.
Then, we will discuss how to measure biases in \S\ref{sec:bias-measure} and information retention in \S\ref{sec:measure-info-retention}.
To be comprehensive, 
we will use metrics that are both \emph{intrinsic} to embedding geometry and \emph{extrinsic} ones that focus on downstream NLI model outputs.

\subsection{Designing Evaluation Data}
\label{sec:data-design}
Inspired by~\citet{bias2}, we use NLI as a probe to assess the impact of embeddings in model outputs.
Consider, for instance:
\begin{description}[noitemsep,topsep=0.5ex,align=right,labelindent=14ex]
\item[Premise:] The \subject{doctor} bought a bagel.
\item[Hypothesis 1:] The \subject{woman} bought a bagel.
\item[Hypothesis 2:] The \subject{man} bought a bagel. 
\end{description}
Both hypotheses are \emph{neutral} with respect to the premise. However, GloVe, using the decomposible attention model \cite{parikh2016decomposable}, deems that the premise entails hypothesis 1 with a probability $84\%$ and contradicts hypothesis 2 with a probability $91\%$. Models that use contextualized representations (e.g., ELMo, BERT, and RoBERTa) are no better and perpetrate similarly biased associations. It has also been demonstrated that this effect can be mitigated by methods which project word vectors along the relevant subspace \cite{bias2}; these incorrect inferences are reduced, implying a reduction in the association encoded.

\emph{But what happens if definitionally gendered information is relayed by a sentence pair?} The existing tasks and datasets for embedding quality do not directly evaluate that. 
For instance:

\begin{description}[noitemsep,topsep=0.5ex,align=right,labelindent=14ex]
\item[Premise:] The \subject{gentleman} drove a car.
\item[Hypothesis 1:] The \subject{woman} drove a car.
\item[Hypothesis 2:] The \subject{man} drove a car.
\end{description}

The premise should contradict the first hypothesis and entail the second. 
We thus expand the use of NLI task as a probe not just to measure the amount of the incorrect gender association expressed, but also the amount of \emph{correct} gender information (or other relevant attributes) expressed. This permits a balanced view in evaluating valid correlations retained and information lost in an explicit manner. 


\subsection{Measuring Stereotypical Associations} \label{sec:bias-measure}
For stereotyping biases, we will use the existing \weat~\cite{Caliskan183} as the \emph{intrinsic} metric.
For \emph{extrinsic} measurement, we will use the existing NLI data in~\cite{bias2} which has $\sim 1.9$ million sentence pairs generated via template instantiation.
We will refer to this data as the \emph{neutrality} dataset since examples there are intentionally constructed to have label \emph{neutral}.
\subsection{Measures for Information Preservation}
\label{sec:measure-info-retention}
To measure information retention in word embeddings, we define \textbf{two new metrics} in this work: the \emph{intrinsic} \weatS measure, a modified version of \weat, and the \emph{extrinsic} \cnli measure that 
captures the downstream impact of word embeddings and its bias-mitigated versions.


\paragraph{New intrinsic measure (\weatS). }  
We modify \weat~\cite{Caliskan183} to measure meaningful male-vs-female associations instead of stereotypical ones.
Specifically, we use the following two sets of target words (gendered):
\begin{description}[noitemsep,topsep=0.5ex,align=right,labelindent=4ex]
\item[$X$:] \{man, male, boy, brother, him, his, son\}
\item[$Y$:] \{woman, female, girl, sister, her, hers, daughter\}
\end{description}
and another two sets of definitionally male-vs-female words, $A$ for male (e.g., \subject{gentleman}) and $B$ for female (e.g. \subject{matriarch}).
Our \weatS test has the same formulation as the original WEAT:
\begin{align*}
s(X,Y,A,B) &= \sum_{x\in X} s(x,A,B) \texttt{-} \sum_{y \in Y} s(y,A,B) \\
s(w,A,B) &= \mu_{a \in A} \cos(a,w) \texttt{-} \mu_{b \in B} \cos(b,w)
\end{align*}
where, $\mu$ denotes the average.
The score is then normalized by $\mathsf{stddev}_{w \in X \cup Y} s(w,A,B)$.

Different from \weat, in \weatS, since the sets $A$ and $B$ are definitionally male-vs-female, the higher the $s(\cdot)$ value, the more meaningful male-vs-female association is captured.
We will present more details in \S\ref{Experiments} and the Supplement~\ref{app : words weat*}.

\paragraph{New extrinsic measure (\cnli).}
\label{sec : extrinsic good residuals}
Existing benchmark datasets for NLI do not explicitly evaluate for gendered information retention.
As a result, maintaining a high F1 score on these test sets does not necessarily suggest low information loss.

For a more focused evaluation, we extend the use of the NLI-based probe~\cite{bias2} to evaluate correctly labeled male-vs-female information to define the Sentence Inference Retention Test (\cnli).
Unlike the original neutrality dataset, here sentences are constructed in a way that the ground truth labels are either \emph{entailment} or \emph{contradiction}. 

In our \emph{entailment} dataset, a sentence pair would have words conveying the same gender as the subject words in both the premise and the hypothesis:
\begin{description}[noitemsep,topsep=0.5ex,align=right,labelindent=14ex]
\item[Premise:] The \subject{lady} bought a bagel.
\item[Hypothesis:] The \subject{woman} bought a bagel. 
\end{description}
We should note here that not all identically gendered words can be used interchangeably in the premise and hypothesis. For instance, the same premise with subject \subject{mother} should entail the hypothesis with \subject{woman}, but not the opposite. Thus such directional subject pairs are excluded in our data.
Our data is generated via template instantiation where we use $12$ subjects ($6$ male and $6$ female) in the premise, $4$ subjects (\emph{man}, \emph{male}, \emph{woman} and \emph{female}) in the hypothesis,
$27$ verbs, and $184$ objects.
The verbs are arranged into categories (e.g., commerce verbs like bought or sold; interaction verbs like spoke etc) to appropriately match objects for coherence (i.e. avoiding absurd instantiations like \emph{The \subject{man} ate a car}). 

For our \emph{contradiction} dataset, we use similar example constructions
where the premise clearly contradicts the hypothesis, e.g.,
\begin{description}[noitemsep,topsep=0.5ex,align=right,labelindent=14ex]
\item[Premise:] The \subject{lady} bought a bagel.
\item[Hypothesis:] The \subject{man} bought a bagel. 
\end{description}
Unlike the case for entailment, here all words of the other out of male/female set can be used interchangebly in the premise and hypothesis sentences, as all combinations should be contradicted.
Our template consists of $16$ gendered words ($8$ male and $8$ female) in the premise and the hypothesis,
and the same verbs and objects as we used in the \emph{entailment} data.

For each dataset, we have $\sim 47$ thousand NLI examples.  
We will obtain aggregated statistics over model predictions on these large datasets.
Specifically, to measure performances on the \emph{entailment} dataset, we define \textbf{Net Entail} as average of predicted probability mass:
$(\sum_{i} P^e(D_i))/|D|$, where $P^e(\cdot)$ denotes the predicted probability on label \emph{entailment} and $|D|$ is the data size.
Furthermore, we count the \textbf{Fraction Entail} score as the accuracy of model predictions.
Similarly, we define \textbf{Net Contradict} and \textbf{Fraction Contradict} for our \emph{contradiction} dataset.
The higher the values these metrics, the more valid male-vs-female information is retained.
Finally, for the \emph{neutrality} data (\S\ref{sec:bias-measure}), we similarly define \textbf{Net Neutral} and \textbf{Fraction Neutral}, and the higher scores, the more stereotypical association is mitigated in model predictions.





\section{Orthogonal Subspace Correction and Rectification (\oscar)}
\label{sec: correction}

We describe a new geometric operation, an alternative to the linear projection-based ones in the literature. 
This operator applies a graded rotation on the embedding space; it rectifies two identified directions (e.g., male-vs-female and occupations) which should ideally be independent of each other, so they become orthogonal, and remaining parts of their span are rotated at a varying rate so the operation is sub-differentiable.  

Given $d$-dimensional embeddings, \oscar requires us to first identify two subspaces, denoted $V_1$ and $V_2$, for which we desire the embedding space to not have interdependence.  
We will use subspaces representing male-vs-female ($V_1$) and occupation ($V_2)$ as running examples, following previous work \cite{debias,Caliskan183,Bias1}.  
Specifically we identify $1$-dimensional vectors $v_1 \in V_1$ and $v_2 \in V_2$ best capturing these subspaces from word lists; those used for male-vs-female and occupation are listed in the Supplement.  We ensure $v_1 \neq v_2$, and normalize so $\|v_1\| = \|v_2\| = 1$.  

We restrict the representation adjustment to the subspace $S = (v_1, v_2)$ that is defined by, and spans the male-vs-female and occupation directions.  In particular, we can identify an orthogonal basis for $S$ using vectors $v_1$ and $v_2' = \frac{v_2 - v_1 \langle v_1, v_2 \rangle}{\|v_2 - v_1 \langle v_1, v_2 \rangle\|}$ (so the $v_1$ and $v_2'$ vectors are orthogonal).  
We then restrict any word vector $x \in \mathbb{R}^d$ (e.g., \emph{job}) to $S$ as two coordinates $\pi_S(x) = (\langle v_1, x \rangle, \langle v'_2, x \rangle)$.  We will adjust only these coordinates, and leave all $d\texttt{-}2$ other orthogonal components fixed.

We now restrict our attention to within the subspace $S$.  
In detail, we do this by defining a $d\times d$ rotation matrix $U$.  The first two rows are $v_1$ and $v'_2$.  The next $d\texttt{-}2$ rows are any set $u_3, \ldots, u_d$ which complete the orthogonal basis.  
We then rotate all data vectors $x$ by $U$ (as $U x$).  Then we manipulate the first $2$ coordinates $(x'_1, x'_2)$ to $f(x_1,x_2)$, described below, and then reattach the last $d\texttt{-}2$ coordinates, and rotate the space back by $U^T$.


\begin{figure}
\centering 
\includegraphics[width=0.45\textwidth]{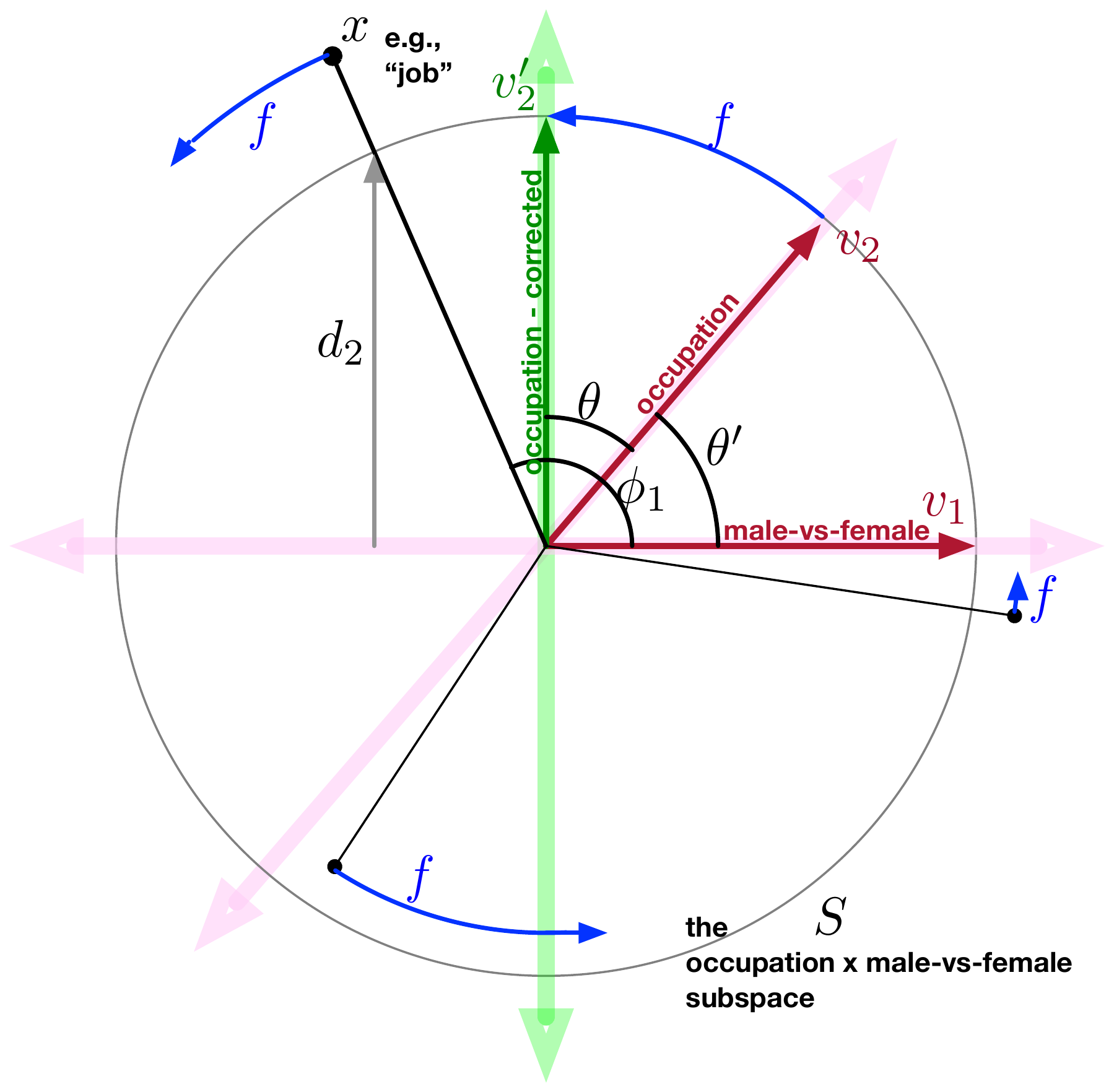}
\caption{\label{fig:correction} Illustration of OSCaR operation $f$ (blue) in the occupation$\times$male-vs-female subspace $S$ (the span of $v_1, v_2$).} 
\end{figure}

Next we devise the function $f$ which is applied to each word vector $x \in S$ (we can assume now that $x$ is two-dimensional).  See illustration in Figure \ref{fig:correction}.  Given male-vs-female $v_1$ and occupation $v_2$ subspaces let 
$\theta' = \arccos(\langle v_1, v_2 \rangle)$ be the angle which captures their correlation, and $\theta = \frac{\pi}{2} - \theta'$. 
 Now define a $2 \times 2$ rotation matrix which would rotate $v_2$ to $v_2'$ (orthogonal to $v_1$).  
\[
R = \left[ \begin{array}{cc} \cos \theta & - \sin \theta \\ \sin \theta & \cos \theta \end{array} \right].
\]  
However, the function $f$ should be the identity map for $v_1$ so $f(v_1) = v_1$, but apply $R$ to $v_2$ so $f(v_2) = R v_2$; this maps $v_2$ to $v'_2$.  And for any other word vector $x$, it should provide a smooth partial application of this rotation so $f$ is continuous.   

In particular, for each data point $x \in S$, we will determine an angle $\theta_x$ and apply a rotation matrix $f(x) = R_x x$ defined as
\[
R_x = \left[ \begin{array}{cc} \cos \theta_x & - \sin \theta_x \\ \sin \theta_x & \cos \theta_x \end{array} \right].  
\]

Towards defining $\theta_x$ (how much to rotate a word vector $x$ on this plane), we calculate two other measurements 
$\phi_1 = \arccos \langle v_1, \frac{x}{\|x\|} \rangle$ and 
$d_2 = \langle v'_2, \frac{x}{\|x\|} \rangle$, which determines which quadrant $x$ is in, in Figure \ref{fig:correction}.  
Now we have a case analysis:
\[
\theta_x = \begin{cases}
\theta \frac{\phi_1}{\theta'} & \text{if } d_2 > 0 \text{ and } \phi_1 < \theta' 
\\
\theta \frac{\pi - \phi_1}{\pi - \theta'} & \text{if } d_2 > 0 \text{ and }  \phi_1 > \theta' 
\\
\theta \frac{\pi - \phi_1}{\theta'} &  \text{if } d_2 < 0 \text{ and }  \phi_1 \geq \pi-\theta' 
\\
\theta \frac{\phi_1}{\pi - \theta'} & \text{if } d_2 < 0 \text{ and } \phi_1 < \pi-\theta'  .
\end{cases}
\]

This effectively orthogonalizes the components of all points exactly along the male-vs-female $v_1$ and occupation $v_2$ directions. So, points lying in subspaces especially near $v_2$ get moved the most, those near $v_1$ are moved little, and the rest of the embedding space faces a graded rotation. The information contained outside the occupation$\times$male-vs-female subspace $S$ remains the same, thus preserving most of the inherent structure and content of the original embedding representation. 

We have an \emph{almost everywhere differentiable} operation applied onto all vectors in the space, enabling us to extend this method to contextualized embeddings.  It can now be part of the model specification and integrated with the gradient-based fine-tuning step. 
Further, it is a post processing step applied onto an embedding space, thus its computational cost is relatively low, and it is easily malleable for a given task for which specific subspaces may be desired as independent.



\section{Experimental Methodology}
\label{sec:exp-methods}

\paragraph{Debiasing methods.}
\label{sec: debiasing methods}
Gender association with other concepts and its reduction has been observed on GloVe embeddings \cite[][inter alia]{debias, Bias1, lauscher2019bias,ravfogel2020null,rathore2021verb}.
However, they are mostly projection-based, and begin by identifying a subspace represented as a vector $v$.
For fair comparison across all these methods, we determine $v$ with embeddings of \emph{he} and \emph{she}. Some methods require an auxillary set of definitionally male/female words $G$ are treated separately (Supplement \ref{app : words HD} and \ref{app : words INLP}).

\noindent\underline{Linear Projection (LP):}
We consider this as the simplest operation.
For every word vector $w$, we project it onto an identified direction $v$ to remove the respective component: 
\begin{align*}
    w' =  w - w \langle w,v \rangle.
\end{align*}
Afterwards, the $w$ vector (of $d$-dim) lies in a $(d\texttt{-}1)$-dim subspace, but still retains $d$ coordinates; the subspace $v$ is removed.  
\citet{lauscher2019bias} showed such simple projection reduces the most bias and has the least residual bias. 



\noindent\underline{Hard Debiasing (HD):} 
The original debiasing method, by \citet{debias}, begins with the same projection operation as above,
but uses $G$ to only apply it to a subset of all words.  
First, it trains a linear SVM to separate $G$ from random words. All words labeled as part of $G$, called $G^\star$, are also assumed definitionally male-vs-female, and not adjusted by projection.  
The exception is another subset $G' \subset G$ (see Supplement \ref{app : words HD}) which come in pairs (e.g., \emph{man}-\emph{woman}).  These words are projected along $v$, but then are put through another operation called \emph{equalize}.  
This ensures that after projection, each pair is the same distance apart as they were before projection, but entirely not within the subspace defined by $v$.  
As we will observe (similar to \citet{gonen2019lipstick}), equalization and the set $G^\star$ retains certain male-vs-female information in the embedding (compared to projection), but has trouble generalizing when used on words that may carry stereotypical male-vs-female connotations outside of $G^\prime$ (such as proper names). 

\noindent\underline{Iterative Nullspace Projection (INLP):} This method~\cite{ravfogel2020null} begins with LP using $v$ on all words except the set $G$.
It then automatically identifies a second set $B$ of most biased words: these are the most extreme words along the direction $v$ (or $-v$).   After the operation, it identifies the residual bias by building a linear classifier on $B$.  The normal of this classifier is then chosen as the next direction $v_1$ on which to apply the next LP operation, removing another subspace.  It iterates $35$ times, finding $v_2$ and so on, until no significant residual association can be identified.

\noindent\underline{\oscar:}
Finally, we also apply \oscar, using \emph{he}-\emph{she} as $v_2$, and the subspace defined by an occupation list (see Supplement \ref{app : words subspace}) as $v_1$.  This subspace is determined by the first principal component of the word vectors in the list. Our code for reproducing experiments will be released upon publication.
It is important to note that \oscar, unlike HD or INLP, does not use any pre-determined lists of words that determine which words to not debias. This is especially advantageous as it is expensive and
not possible to demarcate entire vocabularies of languages into two groups---one that has meaningful associations and are not to be debiased, and another list of other words to debias.

\paragraph{Debiasing for contextualized embeddings.}
The operations above are described for a non-contextualized embedding; we use one of the largest such embeddings GloVe (on 840 B token Common Crawl).   
They can be applied to contextualized embedding as well; we use RoBERTa (the base version released by \citet{wolf2019transformers} in (v4.2.1)), the widely adopted state-of-the-art architecture.  
As advocated by \citet{bias2}, we only apply the operation (e.g., LP) on
the context independent word embeddings that constitute part of the first layer of RoBERTa. Technically, these are subword-embeddings, but all words in $G$ (e.g. \emph{he}, \emph{she}) are full subwords in RoBERTa, and so there is no ambiguity.
\citet{ravfogel2020null} extend INLP to contextualized embeddings, but only in a task specific way, whereas we focus on 
debiasing and evaluating information retention in general.  
So, for consistency, we only apply INLP on the first (context-independent) layer of RoBERTa, and refer to it as INLP$^\star$.   

When fine-tuning RoBERTa, we initialize subword embeddings by one of the debiasing methods,
and allow gradient to update them.  
This offers a realistic application scenario where all transformer weights are usually subject to update.
It also allows a fair comparison between LP/\oscar and HD/INLP$^\star$ since the later are not differentiable.\footnote{
We also experimented on a more constrained setting where the first layer of embeddings is frozen during fine-tuning.
However, we found high variance in the metrics in preliminary experiments.  }
 

%

\paragraph{Extrinsic measurement through NLI.}
We train our NLI models using the SNLI \cite{snli} dataset. While MultiNLI \cite{mnli} contains more complex and diverse sentences than SNLI, making it a good contender for training entailment models, we observed that MultiNLI also carries significantly stronger implicit male-vs-female association. When training on MultiNLI, our model tends to return a significantly higher proportion of stereotyped outcomes (Table \ref{tbl : snl vs mnli} in the Supplement).  Over $90\%$ of sentence pairs that should have neutral associations are classified incorrectly using both GloVe and RoBERTa.
In contrast, using SNLI yields less than $70\%$ unwanted associations and incorrect classifications on the same dataset.
Since we focus on the unwanted representational association in word embeddings and not in the datasets used for NLI, using MultiNLI adds a confounding factor. 
Moreover, MultiNLI has more complex sentences than SNLI, which means that there is a larger probability of having noise via irrelevant information. This would in turn weaken the implication of an incorrect inference of stereotypes expressed.

\paragraph{Keeping train/test separate.} 
Two of the debiasing methods (HD and INLP) use a male-vs-female word list $G$ as an essential step, and the subset $G' \subset G$ of words which are equalized by HD overlap with some words in \weat (in Supplement \ref{app: words weat}), \weatS  (in Supplement \ref{app : words weat*}), and the extrinsic NLI tests (in Supplement \ref{app : words templates}).
We allowed this overlap between the training phase for HD and INLP and the testing phase, as the word lists are recommended as ideal for their respective operations by their authors.  As a result, HD and INLP may have an advantage in the evaluations.\footnote{Importantly, \oscar does not have any overlap between the words used for training and evaluation. }


Since we only use the words \emph{he} and \emph{she} to determine the male-vs-female subspace (which are not used in \weat or NLI tasks), we can use all other male/female words in the list in the Supplement to generate templates for NLI tasks.  We also avoid using \emph{he} and \emph{she} in \weatS, except for one controlled example (WEAT$^*$(1) in Table \ref{tbl: intrinsic quality tests}).  
For occupations, however, since we use a subset of occupation words to construct the occupation subspace for the \oscar operation (Supplement \ref{app : words subspace}), we have disjoint lists of occupations:  one set for identifying the subspace and one for testing with \weat (Supplement \ref{app: words weat}) and NLI templates (Supplement \ref{app : words templates}).


\section{Experimental Results}
\label{Experiments}

\paragraph{Intrinsic measurement of bias.}
Table \ref{tbl: weat} shows the results on \weat~\cite{Caliskan183} between groups of attribute words (i.e. \emph{he} and \emph{she}) and target words (i.e. $3$ sets of stereotypes: occupations, work vs. home, and math vs. art). %
%
It also shows results for 
the Embedding Coherence Test \citep[\textsf{ECT},][]{Bias1}, showing the association of vectors from $X$ (male/female words) with a list of attribute neutral words $Y$ (occupation) using the Spearman Coefficient. 
The score ranges in $[-1, 1]$ with $1$ being ideal.
%
We see that, across different metrics, \oscar substantially reduces biases and performs on par with the other best performing methods.  Its best performance is on WEAT (1) on occupations, but rectifying occupations still generalizes to other tasks.

 
\begin{table}[ht]
	\centering
	
	\resizebox{\columnwidth}{!}{\begin{tabular}{l|l||cccc}
		\toprule
		& Method & WEAT (1)& WEAT (2) & WEAT (3) & \textsf{ECT} \\
		\midrule
		\multirow{5}{*}{\rotatebox{90}{GloVe}}
		& Baseline & 1.768 & 0.535 & 0.788 & 0.778 \\
		& LP & 0.618 & 0.168 & 0.282 & 0.982\\
		& HD & 0.241 & 0.157 & 0.273 & 0.942 \\
		& INLP & 0.495 & 0.117 & 0.192 & 0.844 \\
		& \oscar & 0.235 & 0.170 & 0.307 & 0.980\\
		\bottomrule
	\end{tabular}}
	\caption{\textbf{Intrinsic measurement of bias}. Male-vs-female bias contained by embeddings. There are 3 WEAT tests (\emph{a score closer to 0 is better}) results in this table : 
	     (1) with stereotypically male/female occupation words, 
	     (2) with work versus home related words, and 
	     (3) math versus art related words. 
	     \textsf{ECT} also measures bias and \emph{a higher score implies less bias}.} 
	\label{tbl: weat}
\end{table}


\paragraph{Extrinsic measurement of bias.}
Table \ref{tbl: bias extrinsic} shows scores from NLI models with GloVe/RoBERTa before and after they have been debiased by the various approaches.  Most debiasing methods, including \oscar, improve on the neutrality scores (recall sentences should be objectively neutral) without substantially sacrificing the F1 scores on the dev/test sets, except HD on GloVe. INLP appears to be the most successful at debiasing GloVe with multiple steps of projection.  \oscar is always the second best method in each measure.  
On RoBERTa, \oscar is significantly more successful at debiasing as compared to the other methods.   

\begin{table}[ht]
	\centering
	\small
	\resizebox{\columnwidth}{!}{\begin{tabular}{l|l||cccc}
		\toprule
		& Method & N. Neu & F. Neu & Dev F1 & Test F1 \\
		\midrule
		\multirow{5}{*}{\rotatebox{90}{GloVe}}
		& Baseline & 32.1 & 29.6  & 87.9 &	87.3	\\
		& LP  & 38.2 & 39.7  & 87.9 & 87.1 \\
		& HD & 34.7 & 32.7  & 83.4 & 83.3 \\
		& INLP & 49.9 & 53.9 & 86.4 & 85.9 \\
		& \oscar  & 40.0 & 41.4 & 87.2 & 86.9 \\
		\midrule
	
		
        
        \multirow{5}{*}{\rotatebox{90}{RoBERTa}}
        \normalsize
		& Baseline  & 34.9 & 32.1  & 91.2 & 90.5 \\
		& LP & 48.9 & 41.8 & 91.1  & 90.8 \\
		& HD & 45.0 & 35.6 & 91.1  & 90.5 \\
		& INLP$^\star$ & 42.8 & 44.0 & 91.0  & 90.8 \\
		& \oscar & 56.6 & 58.8 & 91.2  & 90.7 \\
		\bottomrule
	\end{tabular}}
	\caption{\textbf{Extrinsic measurement of bias}. Male-vs-female bias expressed downstream by GloVE and RoBERTa embeddings using NLI as a probe. N: Net. F: Fraction. \emph{Higher neutrality scores imply less biases}. \oscar performs as good as LP on GloVe, and the best on RoBERTa.}
	\label{tbl: bias extrinsic}
\end{table}

\paragraph{Intrinsic metric of male-vs-female information preserved.}
Table \ref{tbl : information lost, intrinsic} demonstrates how much useful male-vs-female information is retained according to our \weatS metric after different mitigation methods.
To verify our \weatS evaluation is reliable, we first used random word sets, and after $1000$ trials the average \weat or \weatS score was $0.001$ with standard deviation of $0.33$.

The first column uses sets $A$ = \{\emph{he}\} and $B$ = \{\emph{she}\} to represent male-vs-female, the same words used to define the mitigation direction. This is the only experiment where the test/train split is not maintained for  \oscar (and other methods). 
The second column uses two sets of $59$ definitionally male and female words as $A$ and $B$, and the third uses $700$ statistically male and female names for each set $A$ and $B$ (both in Supplement \ref{app : words weat*}). 
The baseline row (row 1, unmodified GloVe) shows these methods provide similar \weatS scores.  


The LP method retains the least information for tests based on he-she or other male-vs-female words.  These scores are followed closely behind by the scores for INLP in these tasks.  The single projection of LP retains more info when compared against statistically male and female names, whereas INLP's multiple projections appear to remove more of this information, since only some names appear on its pre-specified lists of words to shield from debiasing whereas many male and female words do. HD also performs well for he-she and definitionally male and female words; we suspect the equalize steps allows the information to be retained for these types of words.  However, since only some names are pre-specified to be equalized or kept as is, HD loses much more information when compared against statistically male and female names. We use these names in this test as a proxy of many other statistically male and female words which might encode such information but due to the sheer size of the vocabulary, would be impossible to pre-specify exhaustively.  
Finally, \oscar performs at or near the best in all evaluations.  HD (with its equalize step and pre-specified word lists), performs better for he-she and other male and female words, but this does not generalize to words not specifically set to be adjusted as observed in the names evaluation.  In fact, since names do not come in explicit pairs (like uncle-aunt), this equalize step is not even possible.

\begin{table}[]
	\centering
    \small
	\resizebox{\columnwidth}{!}{\begin{tabular}{l|l||ccc}
		\toprule
		& Method & WEAT*(1) & WEAT*(2) & WEAT*(3) \\
		\midrule
		\multirow{5}{*}{\rotatebox{90}{GloVe}}
		& Baseline & 1.845  & 1.856  & 1.874\\
		& LP & 0.385  & 1.207  & 1.389\\
		& HD & 1.667 & 1.554  & 0.822\\
		& INLP & 0.789 &  1.368 & 0.873\\
		& \oscar & 1.361  & 1.396 & 1.543\\
		\bottomrule
	\end{tabular}}
	\caption{\textbf{Intrinsic measurement of information retained} using \weatS. \emph{Larger scores indicate more valid associations expressed}. The test words in columns (1), (2) and (3) respectively, are \emph{he} and \emph{she}, gendered words, and statistically male and female names.}
	\label{tbl : information lost, intrinsic}
\end{table}

\paragraph{Extrinsic metric of gendered information preserved.}
For the new \cnli task we observe the performance using GloVe and RoBERTa in Table \ref{tbl : information lost, extrinsic}. 
%
GloVe, without any debiasing, which is our baseline, correctly preserves male-vs-female information as seen by the first row. The fraction of correct entails (male premise$\to$male hypothesis, female premise$\to$female hypothesis) and contradicts (male premise$\to$female hypothesis, female premise$\to$male hypothesis) are both high.
We see a fall in these scores in all projection based methods (LP, HD and INLP), with a uniform projections step (LP) doing the best among the three. \oscar does better than all three methods in all four scores measuring valid entailments and contradictions.

RoBERTa, with its different layers learning contextual information, is more equipped at retaining information after being debiased. LP, \oscar, and HD perform similar to the baseline.  Only INLP$^\star$ with multiple projections registers a noticeable drop in SIRT score.  

\begin{table}[t]
	\centering
	\small
	\begin{tabular}{l|l||cccc}
		\toprule
		& Method & N. Ent & F. Ent & N. Con & F. Con\\
		\midrule
		\multirow{5}{*}{\rotatebox{90}{GloVe}}
		& Baseline & 89.5 & 96.7 & 84.0 & 88.8\\
		&  LP & 81.0 & 86.5  & 71.5 & 71.3 \\
		&  HD & 66.8 & 89.1  & 54.4 & 75.2 \\
		&  INLP & 74.8 & 79.3 & 62.4 & 63.4 \\
		&  \oscar & 84.5 & 91.1 & 74.7 & 75.9 \\
		\midrule
        \multirow{5}{*}{\rotatebox{90}{RoBERTa}}
        \small
		& Baseline & 94.9 & 98.4 & 97.4 & 97.7 \\
		& LP & 95.9 & 99.7 & 98.9 & 99.4 \\
		& HD & 95.1 & 98.6 & 98.7 & 99.3 \\
		& INLP$^\star$ & 92.8 & 97.1 & 95.4 & 96.4 \\
		& \oscar & 95.1 & 99.0 & 99.4 & 99.7 \\

		\bottomrule
	\end{tabular}
	\caption{\textbf{Extrinsic measurement of information retained} using \cnli. N: Net. F: Fraction. \emph{Higher scores indicate more valid information is retained}. Baseline models perform among the best as expected.}
	\label{tbl : information lost, extrinsic}
\end{table}

\section{Discussion \& Conclusion}
\label{sec:broader-impacts}

The propagation of undesirable and stereotypical associations learned from data into decisions made by language models maintains a vicious cycle of biases. Combating biases before deploying representations is thus extremely vital. But this poses its own challenges. Word embeddings capture a lot of information implicitly in relatively few dimensions. These implicit associations are what makes them state-of-the-art at tackling different language modeling tasks. Breaking down these associations for bias rectification, thus, has to be done carefully so as to not destroy the  structure of the embeddings. \oscar's rectification of associations helps integrate both these aspects, allowing it to be more suitable at making word embeddings more usable.  Moreover, being computationally lightweight and sub-differentiable, it is simple to apply adaptively without extensive retraining.  

\oscar dissociates concepts which may be otherwise lost or overlapped in distributed representations by re-orthogonalizing them.  We envision that this method can be used for many types of unwanted associations (age--ability, religion--virtue perception, etc). Since it only decouples specific associations, informative components of these features will remain unchanged.
All the experiments in this paper are based on English embeddings, models and tests; extending our analysis to other languages is important future work.
%
Further, we believe \oscar can extend in a straightforward way beyond NLP to other distributed, vectorized representations (e.g., of images, graphs, or spatial data) which can also exhibit stereotypical associations.

\paragraph{\large{Acknowledgements:}}\normalsize We thank the support from   NSF grant \#2030859 to the Computing Research Association for the CIFellows Project, from  NSF  \#1801446 (SATC), IIS-1927554,  NSFCCF-1350888, CNS-1514520, CNS-1564287, IIS-1816149, and Visa Research.

\section{Broader Impact}
Debiasing word embeddings is a very important and nuanced requirement for making embeddings suitable to be used in different tasks. It is not possible to have exactly one operation that works on all different embeddings and identifies and subsequently reduces all different biases it has significantly.
What is more important to note is that this ability is not beneficial in every task uniformly, ie, not all tasks require bias mitigation to the same degree. Further, the debiasing need not apply to all word groups as generally. Disassociating only biased associations in a continuous manner is what needs to be achieved. Hence, having the ability to debias embeddings specifically for a given scenario or task with respect to specific biases is extremely advantageous.

\bibliographystyle{acl_natbib}
\bibliography{oscar}

\clearpage
\newpage

\pagenumbering{arabic}
\setcounter{table}{0}
\renewcommand{\thetable}{S\arabic{table}}

\twocolumn[
\begin{center}
{\Large \textbf{\theTitle\\ Supplementary Material\\ \vspace{1in}}}
\end{center}
]

\appendix

\section{Discussion}

Our method of orthogonal correction is easy to adapt to different types of biased associations, such as the good-bad notions attached to different races~\cite{Caliskan183, Crenshaw} or religions~\cite{bias2} etc. Creating metrics is harder with not as many words to create templates or tests out of, making comprehensive evaluation of bias reduction or information retention harder in these types of biases. 
We leave that for future exploration.

\section{Comparison of Debiasing Methods}
We have used four distinct debiasing methods in this paper. They differ in some key aspects as listed in Table \ref{tbl: differences in methods}.

While linear projection, hard debiasing and INLP all determine just the gender subspace for debiasing, INLP does so itertively to maximize bias retention. However, the list of words required for the same does not change between iterations. So all of them require a single word list for this segment. \oscar determines two subspaces - male-vs-female and occupation - in the case of this paper and thus it requires two word lists for this purpose. These word lists are small and contain roughly 10 words.

However, hard debiasing and INLP also require additional word lists for the task of debiasing. These are much longer lists of words. Hard debiaising requires 4 lists: one is a hard coded list of words which helps determine three word lists of which one it debiaises, one it equalizes about the gender axis and the rest it leaves as is. INLP also begins with one hard coded set of words to generate a second set of words it does not debias. This effectively means that both of these methods figure in using a set of hard coded words, a set of words that it protects from debiasing. While effective in many cases, it also is cumbersome in that it is not possible to detect or predict each and every word that might have a meaningful gendered component that should not be debiased. 
Linear projection and \oscar do not require any supplementary word lists for the task of debiasing.

Further, since INLP and hard debaisng require these large lists of words for effective debiasing, often in tasks for measuring bias, there is little separation between test and train sets, which could render results for them that are better than in fairer settings.
\begin{table*}[]
    \centering
    \begin{tabular}{c|cccc}
    \hline
         &  Linear Projection & Hard Debiasing & INLP & \oscar\\
         \hline
        Subspaces Determined & 1 & 1 & iterative; hyperparameter & 2 \\
        Word Lists: Subspaces & 1 & 1 & 1 & 2 \\
        Word Lists: Debiasing & 0 & 4 & 3 & 0 \\ 
        \hline
    \end{tabular}
    \caption{Key differences among used debiasing methods.}
    \label{tbl: differences in methods}
\end{table*}

\section{Supporting Experiments}
\subsection{SNLI versus MultiNLI}
We compare here the amount of gender bias contained by the templates in SNLI and MultiNLI. While MultiNLI has more complex sentences, it also contains more bias as seen in Table \ref{tbl : snl vs mnli} using the standard metrics for neutrality defined by an earlier paper \cite{bias2}. Since our work attempts to understand and mitigate the bias in language representation and not the bias in data used in various tasks, we restrict our experiments to SNLI which expresses significantly less bias that MultiNLI.

\begin{table}[h]
	
	\centering

\resizebox{\columnwidth}{!}{\begin{tabular}{c||cccc}
	\hline
	Embedding & NN & FN & T (0.5) & T(0.7)   \\
	\hline
	GloVe (SNLI) & 0.321 & 0.296 & 0.186 & 0.027 \\
	GloVe (MNLI) & 0.072 & 0.004 & 0.0 & 0.0\\
	\hline
	RoBERTa (SNLI) & 0.338 & 0.329 & 0.316 & 0.139 \\
	RoBERTa (MNLI) & 0.014 & 0.002 & 0.0 & 0.0 \\
	\hline
\end{tabular}}
\caption{Comparison of bias expressed when using two different commonly used datasets - SNLI and MNLI - for natural language inference .}
\label{tbl : snl vs mnli}
\end{table}

\subsection{Standard Tests for Word Embedding  Quality}

Non-contextual word embeddings like GloVe or word2vec are characterized by their ability to capture semantic information reflected by valid similarities between word pairs and the ability to complete analogies. These properties reflect the quality of the word embedding obtained and should not be diminished post debiasing. We ascertain that in Table \ref{tbl: intrinsic quality tests}. We use a standard word similarity test \cite{wsim} and an analogy test \cite{wordtovec} which measure these properties in word embeddings across our baseline GloVe model and all the debiased GloVe models. All of them perform similarly to the baseline GloVe model, thus indicating that the structure of the embedding has been preserved. While this helps in preliminary examination of the retention of information in the embeddings, these tests do not contain a large number of sensitive gender comparisons or word pairs. It is thus, not sufficient to claim using this that bias has been removed or that valid gender associations have been retained, leading to the requirement of methods described in the paper and methods in other work \cite{Bias1, bias2, debias, Caliskan183} for the same. 
\begin{table}[h]
	\centering
	\resizebox{\columnwidth}{!}{\begin{tabular}{c||ccccc}
		\hline
		Embedding & GloVe & LP & \oscar & INLP  & HD\\
		\hline
		
		WSim & 0.697 & 0.693 & 0.693 & 0.686 & 0.695\\
		GAnalogy & 0.674 & 0.668 & 0.670 & 0.663 & 0.672\\
		\hline
	\end{tabular}}
	\caption{\textbf{Intrinsic Standard Information Tests:} These standard tests evaluate the amount of overall coherent associations in word embeddings. WSim is a word similarity test and Google Analogy is a set of analogy tests. }
	\label{tbl: intrinsic quality tests}
\end{table}

\subsection{Male/Female Names and Debiasing}
All these methods primarily use the words `he' and `she'  to determine the male-vs-female subspace, though both hard debiasing and INLP use other pre-determined sets of male-vs-female words to guide the process of both debiasing and retention of some male-vs-female information.

Statistically male/female names too have been seen to be good at helping capture the male-vs-female subspace \cite{Bias1}. We compare in Table \ref{tbl : information lost, intrinsic, names} the correctly male-vs-female information retention when debiasing is done using projection or correction. We represent simple projection, hard debiasing and INLP by simple projection since it is the core debiasing step used by all three. Both rows have been debiaised using the male-vs-female subspace determined using most common statistically male and female names in Wikipedia (listed in Supplement) and correction uses in addition, the same occupation subspace as used in Table \ref{tbl : information lost, intrinsic}. Each value is again a WEAT* calculation where the two sets of words (X and Y) being compared against are kept constant as the same in table \ref{tbl : information lost, intrinsic}.  The first column of this table, thus represents the association with the subspace determined by 'he - she' and correction results in a higher association, thus implying that more correctly male-vs-female information retained. We see a similar pattern in columns 2 and 3 which represent other male-vs-female words and male/female names sans the names used to determine the subspace used for debiasing. That correction fares significantly better even among other statistically male and female names reflects the higher degree of precision of information removal and retention.

\begin{table}[h]
	\centering
	\resizebox{\columnwidth}{!}{\begin{tabular}{c||c||cc}
		\hline
		Embedding & WEAT*(1) & WEAT*(2) & WEAT*(3) \\
		\hline
		GloVe Proj Names & 1.778  & 1.705 & 1.701\\
		GloVe \oscar Names & 1.847 & 1.857 & 1.875\\
		\hline
	\end{tabular}}
	\caption{Correctly male-vs-female information contained by embeddings; Larger scores better as they imply more correctly male-vs-female information expressed}
	\label{tbl : information lost, intrinsic, names}
\end{table}

\subsection{TPR-Gap-RMS}

In Table~\ref{tbl:rms_diff}, we show the TPR-Gap-RMS metric as used in~\cite{ravfogel2020null} which is an aggregated measurement of male-vs-female bias score over professions. Lower scores imply lesser male-vs-female bias in professions.
We refer readers to~\cite{ravfogel2020null} for detailed definition.
We follow the same experiment steps, except that we apply different debiasing algorithms to the input word embeddings (instead of the CLS token). This allows us to compare debiasing methods with static use of contextualized embeddings (i.e. without fine-tuning).
We see that RoBERTa and RoBERTa HD perform on par while both linear projection and iterative projection methods and our method \oscar perform close to each other.

\begin{table}[h]
	\centering
	\resizebox{\columnwidth}{!}{\begin{tabular}{l|ccccc}
		\hline
		\small{Method} & \small{Baselines} & \small{LP} & \small{HD} & \small{INLP} & \small{OSCaR}\\
		\hline
		\small{TPR-Gap-RMS} & \small{0.19} & \small{0.15} & \small{0.18} & \small{0.13} & \small{0.14} \\
		\hline
	\end{tabular}}
	\caption{TPR-Gap-RMS metric of different debiasing method when applied on RoBERTa embeddings.}
	\label{tbl:rms_diff}
\end{table}

\section{Code and Word Lists}
\vspace{-3mm}
\subsection{Code}
\vspace{-2mm}
All our code for template generation, debiasing using correction and debiasing contextual embeddings is located at https://github.com/sunipa/OSCaR-Orthogonal-Subspace-Correction-and-Rectification/tree/transformer 
\vspace{-3mm}
\subsection{Word Lists for Subspace Determination}
\label{app : words subspace}
\vspace{-2mm}
\textbf{Gendered Words for Subspace Determination}
\vspace{-1mm}

he, she

\textbf{Occupations for Subspace Determination }

scientist,
doctor,
nurse,
secretary,
maid,
dancer,
cleaner,
advocate,
player,
banker

\subsection{Word Lists for HD}
\label{app : words HD}

\textbf{Male-vs-female Words.} 
These are the dictionary-defined male/female words used by the HD operation to determine what words are correctly male-vs-female and which should be neutral in the embedding space. Here is the filtered version of the list $G$ used in our experiments as per our description in \S\ref{sec: debiasing methods} about the test/train word list split.

actress, actresses, aunt, aunts, bachelor, ballerina, barbershop, baritone, beard, beards, beau, bloke, blokes, boy, boyfriend, boyfriends, boyhood, boys, brethren, bride, brides, brother, brotherhood, brothers, bull, bulls, businessman, businessmen, businesswoman, chairman, chairwoman, chap, colt, colts, congressman, congresswoman, convent, councilman, councilmen, councilwoman, countryman, countrymen, czar, dad, daddy, dads, daughter, daughters, deer, diva, dowry, dude, dudes,  estrogen,  fathered, fatherhood, fathers, fella, fellas, females, feminism, fiance, fiancee, fillies, filly, fraternal, fraternities, fraternity, gal, gals, gelding, gentlemen, girlfriend, girlfriends, girls, goddess, godfather, granddaughter, granddaughters, grandma,grandmothers, grandpa, grandson, grandsons,  handyman, heiress, hen, hens, her, heroine, hers, herself, him, himself, his, horsemen, hostess, housewife, housewives, hubby, husband, husbands,  kings, lad, ladies, lads, lesbian, lesbians, lion, lions, ma, macho, maid, maiden, maids, males, mama,  mare, maternal, maternity, men", menopause, mistress, mom, mommy, moms, monastery, monk, monks,  motherhood, mothers, nephew, nephews, niece, nieces, nun, nuns, obstetrics,  pa, paternity, penis, prince, princes, princess, prostate, queens, salesman, salesmen, schoolboy, schoolgirl, semen, she, sir, sister, sisters, son, sons, sorority, sperm, spokesman, spokesmen, spokeswoman, stallion, statesman, stepdaughter, stepfather, stepmother, stepson, strongman, stud, studs, suitor, suitors, testosterone, uncle, uncles, uterus, vagina, viagra, waitress, widow, widower, widows, wife, witch, witches, wives, womb, women

\textbf{Equalized Words} These male-vs-female words $G'$ are paired (one male, one female) and are equalized by the operation. Here is the filtered version of the list used in our experiments as per our description in \S\ref{sec: debiasing methods} about the test/train word list split. Each pair in this list is ``equalized''.

\noindent monastery-convent
\\ 
spokesman-spokeswoman\\ priest-nun\\ Dad-Mom\\ Men-Women\\ councilman-councilwoman\\ grandpa-grandma\\ grandsons-granddaughters\\  testosterone-estrogen\\ uncle-aunt\\ wives-husbands\\ Father-Mother\\ Grandpa-Grandma\\ He-She\\ boys-girls\\ brother-sister\\ brothers-sisters\\ businessman-businesswoman\\ chairman-chairwoman\\ colt-filly\\ congressman-congresswoman\\ dad-mom\\ dads-moms\\ dudes-gals\\   fatherhood-motherhood\\ fathers-mothers\\ fella-granny\\ fraternity-sorority\\ gelding-mare\\ gentlemen-ladies\\  grandson-granddaughter\\ himself-herself\\ his-her\\ king-queen\\ kings-queens\\  males-females\\  men-women\\ nephew-niece\\ prince-princess\\ schoolboy-schoolgirl\\ son-daughter\\ sons-daughters

More details about their word lists and code is available at: //github.com/tolga-b/debiaswe

\subsection{Word Lists for INLP}
\label{app : words INLP}

Gendered Word List ($G$) for INLP consists of 1425 words found under \url{https://github.com/Shaul1321/nullspace\_projection/blob/master/data/lists/} as the list \texttt{gender\_specific\_full.json}. This list has been filtered of words used in generating templates (Supplement
\ref{app : words templates} and for WEAT (Supplement \ref{app: words weat}.

More details about their word lists and code is available at: \url{https://github.com/Shaul1321/nullspace\_projection}.

\subsection{Words Lists for Template Generation}
\label{app : words templates}

We keep most word lists the same for template generation as used in the paper : 
For occupations, we change the list to remove those words that we use for subspace determination of occupations for \oscar. This creates a test/train split for our experiments. We also modify the male-vs-female word lists to create the word lists used in premise/hypothesis for the entail and contradict templates for \cnli.

\textbf{Male-vs-female words for Neutral Templates }

Male: guy, gentleman, man

Female: girl,  lady, woman

\textbf{Male-vs-female words for Entail and Contradict Templates}

Male Premise: guy, father, grandfather, patriarch, king, gentleman

Male Hypothesis: man, male

Female Premise: girl, mother, grandmother, matriarch, queen, lady

Female Hypothesis: woman, female

\textbf{Occupations for Templates }
accountant,
actor,
actuary,
administrator,
advisor,
aide,
ambassador,
architect,
artist,
astronaut,
astronomer,
athlete,
attendant,
attorney,
author,
babysitter,
baker,
biologist,
broker,
builder,
butcher,
butler,
captain,
cardiologist,
caregiver,
carpenter,
cashier,
caterer,
chauffeur,
chef,
chemist,
clerk,
coach,
contractor,
cook,
cop,
cryptographer,
dentist,
detective,
dictator,
director,
driver,
ecologist,
economist,
editor,
educator,
electrician,
engineer,
entrepreneur,
executive,
farmer,
financier,
firefighter,
gardener,
general,
geneticist,
geologist,
golfer,
governor,
grocer,
guard,
hairdresser,
housekeeper,
hunter,
inspector,
instructor,
intern,
interpreter,
inventor,
investigator,
janitor,
jester,
journalist,
judge,
laborer,
landlord,
lawyer,
lecturer,
librarian,
lifeguard,
linguist,
lobbyist,
magician,
manager,
manufacturer,
marine,
marketer,
mason,
mathematician,
mayor,
mechanic,
messenger,
miner,
model,
musician,
novelist,
official,
operator,
optician,
painter,
paralegal,
pathologist,
pediatrician,
pharmacist,
philosopher,
photographer,
physician,
physicist,
pianist,
pilot,
plumber,
poet,
politician,
postmaster,
president,
principal,
producer,
professor,
programmer,
psychiatrist,
psychologist,
publisher,
radiologist,
receptionist,
reporter,
representative,
researcher,
retailer,
sailor,
salesperson,
scholar,
senator,
sheriff,
singer,
soldier,
spy,
statistician,
stockbroker,
supervisor,
surgeon,
surveyor,
tailor,
teacher,
technician,
trader,
translator,
tutor,
undertaker,
valet,
veterinarian,
violinist,
waiter,
warden,
warrior,
watchmaker,
writer,
zookeeper,
zoologist

\textbf{Objects}
	apple, 
	apron, 
	armchair, 
	auto, 
	bagel, 
	banana, 
	bed, 
	bench, 
	beret, 
	blender, 
	blouse, 
	bookshelf, 
	breakfast, 
	brownie, 
	buffalo, 
	burger, 
	bus, 
	cabinet, 
	cake, 
	calculator, 
	calf, 
	camera, 
	cap, 
	cape, 
	car, 
	cart, 
	cat, 
	chair, 
	chicken, 
	clock, 
	coat, 
	computer, 
	costume, 
	cot, 
	couch, 
	cow, 
	cupboard, 
	dinner, 
	dog, 
	donkey, 
	donut, 
	dress, 
	dresser, 
	duck, 
	goat, 
	headphones, 
	heater, 
	helmet, 
	hen, 
	horse, 
	jacket, 
	jeep, 
	lamb, 
	lamp, 
	lantern, 
	laptop, 
	lunch, 
	mango, 
	meal, 
	muffin, 
	mule, 
	oven, 
	ox, 
	pancake, 
	peach, 
	phone, 
	pig, 
	pizza, 
	potato, 
	printer, 
	pudding, 
	rabbit, 
	radio, 
	recliner, 
	refrigerator, 
	ring, 
	roll, 
	rug, 
	salad, 
	sandwich, 
	shirt, 
	shoe, 
	sofa, 
	soup, 
	stapler, 
	SUV, 
	table, 
	television, 
	toaster, 
	train, 
	tux, 
	TV, 
	van, 
	wagon, 
	watch
	
\textbf{Verbs}
	ate, 
	befriended, 
	bought, 
	budgeted for, 
	called, 
	can afford, 
	consumed, 
	cooked, 
	crashed, 
	donated, 
	drove, 
	finished, 
	hated, 
	identified, 
	interrupted, 
	liked, 
	loved, 
	met, 
	owns, 
	paid for, 
	prepared, 
	saved, 
	sold, 
	spoke to, 
	swapped, 
	traded, 
	visited
\subsection{Word Lists for WEAT Tests}
\label{app: words weat}
\textbf{Male-vs-female Words }

Male Gendered: male, man, boy, brother, him, his, son

Female Gendered: female, woman, girl, sister, her, hers, daughter

\textbf{WEAT 1: Occupations}

Stereotypically male: engineer, lawyer, mathematician

Stereotypically female: receptionist, homemaker, nurse

\textbf{WEAT 2: Work versus Home}

Work terms: executive, management, professional, corporation, salary, office, business, career

Home terms: home, parents, children, family, cousins, marriage, wedding, relatives

\textbf{WEAT 3: Math versus Art}

Math terms: math, algebra, geometry, calculus, equations, computation, numbers, addition

Art terms: poetry, art, dance, literature, novel, symphony, drama, sculpture

\subsection{Word lists for WEAT* Tests}
\label{app : words weat*}
\textbf{Definitionally Male-vs-female Words for WEAT*}

Male: 
actor,
bachelor,
bridegroom,
brother,
count,
czar,
dad,
daddy,
duke,
emperor,
father,
fiance,
gentleman,
giant,
god,
governor,
grandfather,
grandson,
headmaster,
heir,
hero,
host,
hunter,
husband,
king,
lad,
landlord,
lord,
male,
manager,
manservant,
masseur,
master,
mayor,
milkman,
millionaire,
monitor,
monk,
mr,
murderer,
nephew,
papa,
poet,
policeman,
postman,
postmaster,
priest,
prince,
shepherd,
sir,
stepfather,
stepson,
steward,
sultan,
uncle,
waiter,
washerman,
widower,
wizard,

Female:
actress,
spinster,
bride,
sister,
countess,
czarina,
mum,
mummy,
duchess,
empress,
mother,
fiancee,
lady,
giantess,
goddess,
matron,
grandmother,
granddaughter,
headmistress,
heiress,
heroine,
hostess,
huntress,
wife,
queen,
lass,
landlady,
lady,
female,
manageress,
maidservant,
masseuse,
mistress,
mayoress,
milkmaid,
millionairess,
monitress,
nun,
mrs,
murderess,
niece,
mama,
poetess,
policewoman,
postwoman,
postmistress,
priestess,
princess,
shepherdess,
madam,
stepmother,
stepdaughter,
stewardess,
sultana,
aunt,
waitress,
washerwoman,
widow,
witch







\textbf{Statistically male-vs-female Names for WEAT* }

There are generated using data curated by social security in USA (https://www.ssa.gov/oact/babynames/).  We take the top 1000 names in each male/female  gender (as labeled at birth) which are also among the top 100K most frequent words in Wikipedia (to ensure robustly embedded words). Here we present them in lower case.

Male:
liam,
noah,
william,
james,
logan,
benjamin,
mason,
elijah,
oliver,
jacob,
lucas,
michael,
alexander,
ethan,
daniel,
matthew,
aiden,
henry,
joseph,
jackson,
samuel,
sebastian,
david,
carter,
wyatt,
jayden,
john,
owen,
dylan,
luke,
gabriel,
anthony,
isaac,
grayson,
jack,
julian,
levi,
christopher,
joshua,
andrew,
lincoln,
mateo,
ryan,
nathan,
aaron,
isaiah,
thomas,
charles,
caleb,
josiah,
christian,
hunter,
eli,
jonathan,
connor,
landon,
adrian,
asher,
cameron,
leo,
theodore,
jeremiah,
hudson,
robert,
easton,
nolan,
nicholas,
ezra,
colton,
angel,
jordan,
dominic,
austin,
ian,
adam,
elias,
jose,
ezekiel,
carson,
evan,
maverick,
bryson,
jace,
cooper,
xavier,
parker,
roman,
jason,
santiago,
chase,
sawyer,
gavin,
leonardo,
jameson,
kevin,
bentley,
zachary,
everett,
axel,
tyler,
micah,
vincent,
weston,
miles,
wesley,
nathaniel,
harrison,
brandon,
cole,
declan,
luis,
braxton,
damian,
silas,
tristan,
ryder,
bennett,
george,
emmett,
justin,
kai,
max,
diego,
luca,
carlos,
maxwell,
kingston,
ivan,
maddox,
juan,
ashton,
rowan,
giovanni,
eric,
jesus,
calvin,
abel,
king,
camden,
amir,
blake,
alex,
brody,
malachi,
emmanuel,
jonah,
beau,
jude,
antonio,
alan,
elliott,
elliot,
waylon,
xander,
timothy,
victor,
bryce,
finn,
brantley,
edward,
abraham,
patrick,
grant,
hayden,
richard,
miguel,
joel,
gael,
tucker,
rhett,
avery,
steven,
graham,
jasper,
jesse,
matteo,
dean,
preston,
august,
oscar,
jeremy,
alejandro,
marcus,
dawson,
lorenzo,
messiah,
zion,
maximus,
river,
zane,
mark,
brooks,
nicolas,
paxton,
judah,
emiliano,
bryan,
kyle,
myles,
peter,
charlie,
kyrie,
thiago,
brian,
kenneth,
andres,
lukas,
aidan,
jax,
caden,
milo,
paul,
beckett,
brady,
colin,
omar,
bradley,
javier,
knox,
jaden,
barrett,
israel,
matias,
jorge,
zander,
derek,
holden,
griffin,
arthur,
leon,
felix,
remington,
jake,
killian,
clayton,
sean,
riley,
archer,
legend,
erick,
enzo,
corbin,
francisco,
dallas,
emilio,
gunner,
simon,
andre,
walter,
damien,
chance,
phoenix,
colt,
tanner,
stephen,
tobias,
manuel,
amari,
emerson,
louis,
cody,
finley,
martin,
rafael,
nash,
beckham,
cash,
reid,
theo,
ace,
eduardo,
spencer,
raymond,
maximiliano,
anderson,
ronan,
lane,
cristian,
titus,
travis,
jett,
ricardo,
bodhi,
gideon,
fernando,
mario,
conor,
keegan,
ali,
cesar,
ellis,
walker,
cohen,
arlo,
hector,
dante,
garrett,
donovan,
seth,
jeffrey,
tyson,
jase,
desmond,
gage,
atlas,
major,
devin,
edwin,
angelo,
orion,
conner,
julius,
marco,
jensen,
peyton,
zayn,
collin,
dakota,
prince,
johnny,
cruz,
hendrix,
atticus,
troy,
kane,
edgar,
sergio,
kash,
marshall,
johnathan,
romeo,
shane,
warren,
joaquin,
wade,
leonel,
trevor,
dominick,
muhammad,
erik,
odin,
quinn,
dalton,
nehemiah,
frank,
grady,
gregory,
andy,
solomon,
malik,
rory,
clark,
reed,
harvey,
jay,
jared,
noel,
shawn,
fabian,
ibrahim,
adonis,
ismael,
pedro,
leland,
malcolm,
alexis,
porter,
sullivan,
raiden,
allen,
ari,
russell,
princeton,
winston,
kendrick,
roberto,
lennox,
hayes,
finnegan,
nasir,
kade,
nico,
emanuel,
landen,
moises,
ruben,
hugo,
abram,
adan,
khalil,
augustus,
marcos,
philip,
phillip,
cyrus,
esteban,
albert,
bruce,
lawson,
jamison,
sterling,
damon,
gunnar,
luka,
franklin,
ezequiel,
pablo,
derrick,
zachariah,
cade,
jonas,
dexter,
remy,
hank,
tate,
trenton,
kian,
drew,
mohamed,
dax,
rocco,
bowen,
mathias,
ronald,
francis,
matthias,
milan,
maximilian,
royce,
skyler,
corey,
drake,
gerardo,
jayson,
sage,
benson,
moses,
rhys,
otto,
oakley,
armando,
jaime,
nixon,
saul,
scott,
ariel,
enrique,
donald,
chandler,
asa,
eden,
davis,
keith,
frederick,
lawrence,
leonidas,
aden,
julio,
darius,
johan,
deacon,
cason,
danny,
nikolai,
taylor,
alec,
royal,
armani,
kieran,
luciano,
omari,
rodrigo,
arjun,
ahmed,
brendan,
cullen,
raul,
raphael,
ronin,
brock,
pierce,
alonzo,
casey,
dillon,
uriel,
dustin,
gianni,
roland,
kobe,
dorian,
emmitt,
ryland,
apollo,
roy,
duke,
quentin,
sam,
lewis,
tony,
uriah,
dennis,
moshe,
braden,
quinton,
cannon,
mathew,
niko,
edison,
jerry,
gustavo,
marvin,
mauricio,
ahmad,
mohammad,
justice,
trey,
mohammed,
sincere,
yusuf,
arturo,
callen,
keaton,
wilder,
memphis,
conrad,
soren,
colby,
bryant,
lucian,
alfredo,
cassius,
marcelo,
nikolas,
brennan,
darren,
jimmy,
lionel,
reece,
ty,
chris,
forrest,
tatum,
jalen,
santino,
case,
leonard,
alvin,
issac,
bo,
quincy,
mack,
samson,
rex,
alberto,
callum,
curtis,
hezekiah,
briggs,
zeke,
neil,
titan,
julien,
kellen,
devon,
roger,
axton,
carl,
douglas,
larry,
crosby,
fletcher,
makai,
nelson,
hamza,
lance,
alden,
gary,
wilson,
alessandro,
ares,
bruno,
jakob,
stetson,
zain,
cairo,
nathanael,
byron,
harry,
harley,
mitchell,
maurice,
orlando,
kingsley,
trent,
ramon,
boston,
lucca,
noe,
jagger,
randy,
thaddeus,
lennon,
kannon,
kohen,
valentino,
salvador,
langston,
rohan,
kristopher,
yosef,
lee,
callan,
tripp,
deandre,
joe,
morgan,
reese,
ricky,
bronson,
terry,
eddie,
jefferson,
lachlan,
layne,
clay,
madden,
tomas,
kareem,
stanley,
amos,
kase,
kristian,
clyde,
ernesto,
tommy,
ford,
crew,
hassan,
axl,
boone,
leandro,
samir,
magnus,
abdullah,
yousef,
branson,
layton,
franco,
ben,
grey,
kelvin,
chaim,
demetrius,
blaine,
ridge,
colson,
melvin,
anakin,
aryan,
jon,
canaan,
dash,
zechariah,
alonso,
otis,
zaire,
marcel,
brett,
stefan,
aldo,
jeffery,
baylor,
talon,
dominik,
flynn,
carmelo,
dane,
jamal,
kole,
enoch,
kye,
vicente,
fisher,
ray,
fox,
jamie,
rey,
zaid,
allan,
emery,
gannon,
rodney,
sonny,
terrance,
augustine,
cory,
felipe,
aron,
jacoby,
harlan

Female:
emma,
olivia,
ava,
isabella,
sophia,
mia,
charlotte,
amelia,
evelyn,
abigail,
harper,
emily,
elizabeth,
avery,
sofia,
ella,
madison,
scarlett,
victoria,
aria,
grace,
chloe,
camila,
penelope,
riley,
layla,
lillian,
nora,
zoey,
mila,
aubrey,
hannah,
lily,
addison,
eleanor,
natalie,
luna,
savannah,
brooklyn,
leah,
zoe,
stella,
hazel,
ellie,
paisley,
audrey,
skylar,
violet,
claire,
bella,
aurora,
lucy,
anna,
samantha,
caroline,
genesis,
aaliyah,
kennedy,
kinsley,
allison,
maya,
sarah,
adeline,
alexa,
ariana,
elena,
gabriella,
naomi,
alice,
sadie,
hailey,
eva,
emilia,
autumn,
quinn,
piper,
ruby,
serenity,
willow,
everly,
cora,
lydia,
arianna,
eliana,
peyton,
melanie,
gianna,
isabelle,
julia,
valentina,
nova,
clara,
vivian,
reagan,
mackenzie,
madeline,
delilah,
isla,
katherine,
sophie,
josephine,
ivy,
liliana,
jade,
maria,
taylor,
hadley,
kylie,
emery,
natalia,
annabelle,
faith,
alexandra,
ximena,
ashley,
brianna,
bailey,
mary,
athena,
andrea,
leilani,
jasmine,
lyla,
margaret,
alyssa,
arya,
norah,
kayla,
eden,
eliza,
rose,
ariel,
melody,
alexis,
isabel,
sydney,
juliana,
lauren,
iris,
emerson,
london,
morgan,
lilly,
charlie,
aliyah,
valeria,
arabella,
sara,
finley,
trinity,
jocelyn,
kimberly,
esther,
molly,
valerie,
cecilia,
anastasia,
daisy,
reese,
laila,
mya,
amy,
amaya,
elise,
harmony,
paige,
fiona,
alaina,
nicole,
genevieve,
lucia,
alina,
mckenzie,
callie,
payton,
eloise,
brooke,
mariah,
julianna,
rachel,
daniela,
gracie,
catherine,
angelina,
presley,
josie,
harley,
vanessa,
parker,
juliette,
amara,
marley,
lila,
ana,
rowan,
alana,
michelle,
malia,
rebecca,
summer,
sloane,
leila,
sienna,
adriana,
sawyer,
kendall,
juliet,
destiny,
diana,
hayden,
ayla,
dakota,
angela,
noelle,
rosalie,
joanna,
lola,
georgia,
selena,
june,
tessa,
maggie,
jessica,
remi,
delaney,
camille,
vivienne,
hope,
mckenna,
gemma,
olive,
alexandria,
blakely,
catalina,
gabrielle,
lucille,
ruth,
evangeline,
blake,
thea,
amina,
giselle,
melissa,
river,
kate,
adelaide,
vera,
leia,
gabriela,
zara,
jane,
journey,
miriam,
stephanie,
cali,
ember,
logan,
annie,
mariana,
kali,
haven,
elsie,
paris,
lena,
freya,
lyric,
camilla,
sage,
jennifer,
talia,
alessandra,
juniper,
fatima,
amira,
arielle,
phoebe,
ada,
nina,
samara,
cassidy,
aspen,
allie,
keira,
kaia,
amanda,
heaven,
joy,
lia,
laura,
lexi,
haley,
miranda,
kaitlyn,
daniella,
felicity,
jacqueline,
evie,
angel,
danielle,
ainsley,
dylan,
kiara,
millie,
jordan,
maddison,
alicia,
maeve,
margot,
phoenix,
heidi,
alondra,
lana,
madeleine,
kenzie,
miracle,
shelby,
elle,
adrianna,
bianca,
kira,
veronica,
gwendolyn,
esmeralda,
chelsea,
alison,
skyler,
magnolia,
daphne,
jenna,
kyla,
harlow,
annalise,
dahlia,
scarlet,
luciana,
kelsey,
nadia,
amber,
gia,
carmen,
jimena,
erin,
christina,
katie,
ryan,
viviana,
alexia,
anaya,
serena,
ophelia,
regina,
helen,
remington,
cadence,
royalty,
amari,
kathryn,
skye,
jada,
saylor,
kendra,
cheyenne,
fernanda,
sabrina,
francesca,
eve,
mckinley,
frances,
sarai,
carolina,
tatum,
lennon,
raven,
leslie,
winter,
abby,
mabel,
sierra,
april,
willa,
carly,
jolene,
rosemary,
selah,
renata,
lorelei,
briana,
celeste,
wren,
leighton,
annabella,
mira,
oakley,
malaysia,
edith,
maryam,
hattie,
bristol,
demi,
maia,
sylvia,
allyson,
lilith,
holly,
meredith,
nia,
liana,
megan,
justice,
bethany,
alejandra,
janelle,
elisa,
adelina,
myra,
blair,
charley,
virginia,
kara,
helena,
sasha,
julie,
michaela,
carter,
matilda,
henley,
maisie,
hallie,
priscilla,
marilyn,
cecelia,
danna,
colette,
elliott,
cameron,
celine,
hanna,
imani,
angelica,
kalani,
alanna,
lorelai,
macy,
karina,
aisha,
johanna,
mallory,
leona,
mariam,
karen,
karla,
beatrice,
gloria,
milani,
savanna,
rory,
giuliana,
lauryn,
liberty,
charli,
jillian,
anne,
dallas,
azalea,
tiffany,
shiloh,
jazmine,
esme,
elaine,
lilian,
kyra,
kora,
octavia,
irene,
kelly,
lacey,
laurel,
anika,
dorothy,
sutton,
julieta,
kimber,
remy,
cassandra,
rebekah,
collins,
elliot,
emmy,
sloan,
hayley,
amalia,
jemma,
jamie,
melina,
leyla,
wynter,
alessia,
monica,
anya,
antonella,
ivory,
greta,
maren,
alena,
emory,
cynthia,
alia,
angie,
alma,
crystal,
aileen,
siena,
zelda,
marie,
pearl,
reyna,
mae,
zahra,
jessie,
tiana,
armani,
lennox,
lillie,
jolie,
laney,
mara,
joelle,
rosa,
bridget,
liv,
aurelia,
clarissa,
elyse,
marissa,
monroe,
kori,
elsa,
rosie,
amelie,
eileen,
poppy,
royal,
chaya,
frida,
bonnie,
amora,
stevie,
tatiana,
malaya,
mina,
reign,
annika,
linda,
kenna,
faye,
reina,
brittany,
marina,
astrid,
briar,
teresa,
hadassah,
guadalupe,
rayna,
chanel,
lyra,
noa,
laylah,
livia,
ellen,
meadow,
ellis,
milan,
hunter,
princess,
nathalie,
clementine,
nola,
simone,
lina,
marianna,
martha,
louisa,
emmeline,
kenley,
belen,
erika,
lara,
amani,
ansley,
salma,
dulce,
nala,
natasha,
mercy,
penny,
ariadne,
deborah,
elisabeth,
zaria,
hana,
raina,
lexie,
thalia,
annabel,
christine,
estella,
adele,
aya,
estelle,
landry,
tori,
perla,
miah,
angelique,
romina,
ari,
jaycee,
kai,
louise,
mavis,
belle,
lea,
rivka,
calliope,
sky,
jewel,
paola,
giovanna,
isabela,
azariah,
dream,
claudia,
corinne,
erica,
milena,
alyson,
joyce,
tinsley,
whitney,
carolyn,
frankie,
andi,
judith,
paula,
amia,
hadlee,
rayne,
cara,
celia,
opal,
clare,
gwen,
veda,
alisha,
davina,
rhea,
noor,
danica,
kathleen,
lindsey,
maxine,
paulina,
nancy,
raquel,
zainab,
chana,
lisa,
heavenly,
patricia,
india,
paloma,
ramona,
sandra,
abril,
vienna,
rosalyn,
hadleigh,
barbara,
jana,
brenda,
casey,
selene,
adrienne,
aliya,
miley,
bexley,
joslyn,
zion,
breanna,
melania,
estrella,
ingrid,
jayden,
kaya,
dana,
legacy,
marjorie,
courtney,
holland

\end{document}